\documentclass[runningheads]{llncs}
\usepackage[T1]{fontenc}
\usepackage{graphicx,verbatim}
\usepackage{multirow}
\usepackage{diagbox}
\usepackage{graphicx}
\usepackage{booktabs}
\usepackage{tabularx}
\usepackage{picture}
\usepackage{amsmath}
\usepackage{subcaption}
\usepackage{diagbox} 
\usepackage{cancel}
\usepackage{arydshln}
\usepackage{tabularx}
\usepackage{makecell}
\usepackage{color}
\usepackage{threeparttable}
\usepackage{enumitem}
\usepackage{amssymb}
\usepackage{amsmath}
\usepackage{listings}
\usepackage{xcolor}
\usepackage{textcomp}
\usepackage{marvosym}
\usepackage[colorlinks]{hyperref}
\lstdefinestyle{pythonstyle}{
    language=Python,
    basicstyle=\ttfamily\small,
    breaklines=true,
    commentstyle=\color{green!60!black},
    keywordstyle=\color{blue},
    stringstyle=\color{red},
    numbers=left,
    numberstyle=\tiny\color{gray},
    numbersep=5pt,
    backgroundcolor=\color{gray!10},
    showstringspaces=false,
    frame=single,
    rulecolor=\color{black!30},
    tabsize=4,
    captionpos=t
}
\begin{document}
\title{More performant and scalable: Rethinking contrastive vision-language pre-training of radiology in the LLM era}

\author{Yingtai Li\inst{1,2} 
\and Haoran Lai\inst{1,2} 
\and Xiaoqian Zhou\inst{1,2} 
\and Shuai Ming\inst{3} 
\and Wenxin Ma\inst{1,2} 
\and Wei Wei\inst{3} 
\and Shaohua Kevin Zhou\inst{1,2,4,5} 
$^{\href{mailto:skevinzhou@ustc.edu.cn}{\textrm{\Letter}}}$} 

\authorrunning{Yingtai Li et al.}
\institute{
School of Biomedical Engineering, Division of Life Sciences and Medicine, University of Science and Technology of China (USTC), Hefei Anhui, 230026, China 
\and
Center for Medical Imaging, Robotics, Analytic Computing \& Learning (MIRACLE), Suzhou Institute for Advance Research, USTC, Suzhou Jiangsu, 215123, China
\and
The First Affiliated Hospital of USTC, Division of Life Sciences and Medicine, USTC, Hefei Anhui, 230001, China
\and
Jiangsu Provincial Key Laboratory of Multimodal Digital Twin Technology, Suzhou Jiangsu, 215123, China
\and
State Key Laboratory of Precision and Intelligent Chemistry, USTC, Hefei Anhui 230026, China
\\
    \email{liyingtai@mail.ustc.edu.cn, skevinzhou@ustc.edu.cn}}

\maketitle              

\begin{abstract}

    The emergence of Large Language Models (LLMs) presents unprecedented opportunities to revolutionize medical contrastive vision-language pre-training. 
    In this paper, we show how LLMs can facilitate large-scale supervised pre-training, thereby advancing vision-language alignment. We begin by demonstrate that modern LLMs can automatically extract diagnostic labels from radiology reports with remarkable precision (>96\% AUC in our experiments) without complex prompt engineering, enabling the creation of large-scale "silver-standard" datasets at a minimal cost (~\$3 for 50k CT image-report pairs). Further, we find that vision encoder trained on this "silver-standard" dataset achieves performance comparable to those trained on labels extracted by specialized BERT-based models, thereby democratizing the access to large-scale supervised pre-training. Building on this foundation, we proceed to reveal that supervised pre-training fundamentally improves contrastive vision-language alignment. Our approach achieves state-of-the-art performance using only a 3D ResNet-18 with vanilla CLIP training, including 83.8\% AUC for zero-shot diagnosis on CT-RATE, 77.3\% AUC on RAD-ChestCT, and substantial improvements in cross-modal retrieval (MAP@50=53.7\% for image-image, Recall@100=52.2\% for report-image). These results demonstrate the potential of utilizing LLMs to facilitate {\bf more performant and scalable} medical AI systems. Our code is avaiable at \url{https://github.com/SadVoxel/More-performant-and-scalable}.

\end{abstract}

\section{Introduction}
\label{sec:intro}

The evolution of medical artificial intelligence stands at a critical juncture, where the convergence of large-scale data curation and foundational model development is redefining diagnostic paradigms. 
Computed tomography (CT) imaging has allowed non-invasive high-resolution imaging of anatomical structures, which has been a key part of diagnosis for major diseases. However, the reading of CT images is very time-consuming, usually requires orders of minutes to read compared to X-rays. Besides current burden in reading CT images, the number of CT examinations are still undergoing fast growth \cite{smith2012use}. This escalating demand has fueled intense interest in AI diagnostics, yet existing solutions remain trapped in \textbf{a trilemma of performance, scalability, and development cost}.

Traditional supervised learning approaches, while achieving radiologist-level performance in controlled trials and real-world clinical testing \cite{wang2022development,ardila2019end}, face fundamental scaling limitations due to their reliance on expensive manual annotations. The prohibitive costs associated with curating large-scale annotated datasets for training performant models have driven the field toward self-supervised methods. 
Contrastive Vision-Language Pre-training (CLIP) \cite{radford2021clip} is widely recognized as a potential solution, as its training is based on image-report pairs, which naturally exist in clinical practice. By learning from language supervision, CLIP-style models not only enable open-vocabulary recognition that can potentially generalize to unseen conditions without additional training, but establish more detailed image-to-semantic correspondence, which goes beyond simple one-hot labels to capture detailed descriptions of size, location, and density that better characterize complex medical conditions \cite{shui2025largescale}. This detailed correspondence not only facilitates cross-modal retrieval but also enables multimodal processing with LLMs for more powerful and explainable decision making \cite{alayrac2022flamingo,li2023blip2,liu2024visualinstructtuning}. 

Nevertheless, CLIP-style models still face challenges in achieving effective vision-language alignment, as evidenced by their suboptimal zero-shot performance compared to supervised learning models, which is particularly concerning for medical applications where accuracy is paramount. Despite a few works reporting comparable performance with supervised learning models \cite{zhang2023KAD}, to the best of our knowledge, models that have gone through rigorous real-world clinical testing are still based on supervised learning \cite{wang2022development,ardila2019end,cao2023large,cid2024development,wang2024screening}. 
A potential solution suggest by previous works is to decouple image representation learning with vision-language alignment \cite{zhai2022lit}. Which first learn performant image descriptors with high-quality labels, then refining vision-language alignment in subsequent phase. However, this reintroduces the limitations inherent to supervised learning.

The emergence of Large Language Models (LLMs) presents a new opportunity to address this trilemma. In this work, we show how LLMs can facilitate large-scale supervised pre-training and thus advance vision-language alignment. 
We begin with leveraging the power of LLMs to extract labels from radiology reports. After experimenting with three different LLMs (Deepseek \cite{dai2024deepseekmoe}, Qwen \cite{bai2023qwen} and Doubao \cite{doubao2023}), we demonstrate that LLMs can automatically extract diagnostic labels with exceptional precision without the need for complex prompt engineering. This enables creation of "silver-standard" datasets at unprecedented scale (50k+ CT studies across 18 pathologies). Further, we show that a vision encoder trained on these LLM-extracted labels can achieve performance on par with those using labels that use specialized BERT-based model to obtain, which requires manually annotating thousands of reports to develop. By leveraging LLMs to process reports, we reduce the labelling cost of 50k CT images to only 3 dollars, effectively democratizing access to large-scale supervised pre-training. 

Building on this foundation of strong supervised pre-trained models, we leverage the power of supervised pre-training to advance vision-language alignment. We demonstrate that enhanced visual representations learned from supervised pre-training fundamentally transform contrastive learning dynamics, leading to significantly improved alignment between visual and language modalities.

Our experiments demonstrate that stronger supervised pre-trained models consistently improve zero-shot diagnosis, image-to-image retrieval, and image-to-report cross-modal retrieval. Our framework achieves superior performance using merely 10\% of the training data required by existing state-of-the-art results. 
Using vanilla CLIP training with 3D ResNet-18 as the vision encoder, we are able to achieve an 83.8\% AUC in zero-shot classification, mAP\@50=53.7 and Recall@100=52.2 on the large-scale CT-RATE \cite{hamamci2024ctrate}, outperforming previous state-of-the-art \cite{shui2025largescale,lai2025bridged} by a significant margin. 
For out-of-distribution data, the supervised-pretrained model shows even more significant improvement in generalization, achieving 77.3\% AUC in zero-shot classification on the RAD-chestCT \cite{draelos2021radchestct} dataset. This represents a 7.3\% improvement in absolute AUC over previous state-of-the-art methods with a considerably smaller model.
It is also worth noting that our method is compatible with advanced pipeline design \cite{shui2025largescale,lai2025bridged}, which can be integrated with our approach to push performance even further. 

In summary, we introduce the following contributions: 

\begin{itemize}[label=\textbullet]
    \item We perform the first large-scale evaluation of the quality of diagnostic labels extracted from radiology reports by LLMs
    \item We reveal that a vision encoder trained on LLM-extracted labels can achieve performance comparable to models using labels from high-cost annotation methods, thereby democratizing access to large-scale supervised pre-training
    \item We demonstrate that stronger supervised pre-training continuously improves vision-language alignment, achieving state-of-the-art results in zero-shot diagnosis, image-to-image retrieval, and image-to-report cross-modal retrieval using a 3D ResNet-18 with vanilla CLIP training
\end{itemize}

\section{Method}
\label{sec:method}

\subsection{Scalable LLM-powered label extraction}
We design a simple yet effective prompt template for LLMs to extract abnormality labels from radiology reports, which clearly defines each category and enforces a strict output format (18 comma-separated binary values) in the system prompt and instruct each LLM to classify the presence (1) or absence (0) of 18 specific conditions for a given report. This direct binary classification approach proves easier to implement than previous "summary-query" prompt methods \cite{park2024automated} while still maintaining high precision at a large-scale. 

\begin{figure}[h]
    \centering
    \includegraphics[width=\textwidth]{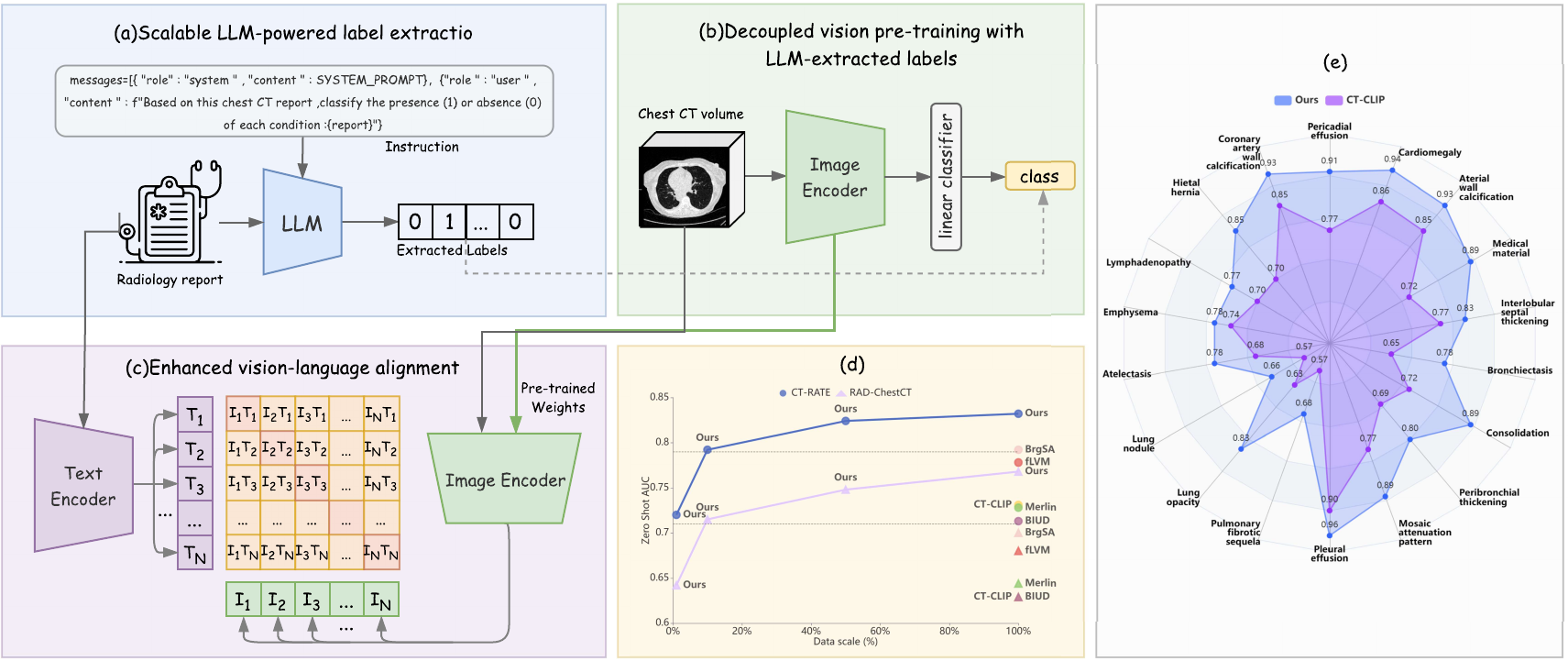}
    \caption{Overview of our proposed framework: (a) LLMs extract binary diagnostic labels from radiology reports, enable curating a large-scale “silver-standard” dataset; (b) these LLM-extracted labels are used for large-scale supervised pre-training to learn high-quality image representations; (c) a decoupled image and text encoder is integrated to align CT images with their corresponding reports; (d) our method surpasses current state-of-the-art approaches using only 10\% of the data, and when trained with 100\% of the data, it significantly outperforms them; and (e) superior performance is achieved across all abnormality classes.}
    \label{fig:overview}
\end{figure}

\subsection{Decoupled vision pre-training with LLM-extracted labels}
Before engaging in vision-language alignment, we first learn superior visual representations by leveraging the LLM-extracted labels. We employ a 3D ResNet-18 model (initialized with kinetics-400 \cite{kay2017kinetics} pre-trained weights from the Torchvision model zoo) as our vision encoder. 
Given an input CT volume $x \in \mathbb{R}^{H \times W \times D}$, our image encoder $f_I$ produces feature maps $F = f_I(x) \in \mathbb{R}^{c \times h \times w \times d}$, where $h,w,d$ are the spatial dimensions of the lowest resolution feature map and $c$ is the number of channels. 
We apply global average pooling to get the features $z = \text{GAP}(F) \in \mathbb{R}^c$, and then employ a linear classifier with sigmoid function:

\begin{equation}
    p_{\text{cls}} = \sigma(W_{\text{cls}}z + b_{\text{cls}}).
\end{equation}

The model is trained using binary cross entropy (BCE) loss:

\begin{equation}
    \mathcal{L}_{\text{BCE}} = -\sum_{i=1}^{N} \left[ y_i \log(p_{\text{cls},i}) + (1-y_i) \log(1-p_{\text{cls},i}) \right].
\end{equation}

We further enhance the classification performance through two additional strategies: (1) Label Smoothing regularization (LS) \cite{szegedy2016labelsmoothing} that prevents the model from becoming over-confident, and (2) Auxiliary Segmentation Supervision (ASS) \cite{li2025supervisionscaling} that leverages automatically generated segmentation masks as additional supervision. These enhanced supervision strategies lead to stronger supervised pre-training models, which in turn improve vision-language alignment.

\subsection{Enhanced vision-language alignment}
Building on the strong visual representations learned via supervised pre-training, we next align vision and language world using a vanilla CLIP training pipeline with image encoder $f_I$ and text encoder $f_T$. Given a batch of $N$ image-text pairs, the encoders produce feature representations $v_i = f_I(I_i) \in \mathbb{R}^{d_I}$ and $t_i = f_T(T_i) \in \mathbb{R}^{d_T}$ respectively. These features are then projected into a shared embedding space, and the alignment is enforced using a contrastive loss:

\begin{equation}
    \mathcal{L}_{\text{CLIP}} = -\frac{1}{2N}\sum_{i=1}^N \left[\log\frac{\exp(z_i^\top t_i)}{\sum_{j=1}^N \exp(z_i^\top t_j)} + \log\frac{\exp(z_i^\top t_i)}{\sum_{j=1}^N \exp(z_j^\top t_i)}\right],
\end{equation}

When using supervised pre-trained models, we explore removing the L2 normalization before computing similarity scores $z_i^\top t_j$ to better preserve the representations learned during the supervised pre-training stage. 
Following CT-CLIP \cite{hamamci2024ctrate,shui2025largescale}, we use CXR-BERT \cite{boecking2022cxrbert} as our text encoder. 
For zero-shot diagnosis evaluation, we follow the protocol of CheXzero \cite{tiu2022chexzero} and CT-CLIP \cite{hamamci2024ctrate}.

\section{Experiment}
\label{sec:experiment}

\subsection{Experiment setup}
\subsubsection{Datasets:}
We use two datasets. \underline{CT-RATE}~\cite{hamamci2024ctrate} contains 50,188 chest CT scans from 21,304 patients with 18 abnormalities identified via BERT-based extraction. We adopt CT-CLIP's splits for training/validation. 
\underline{RAD-ChestCT} \cite{draelos2021radchestct} comprises 36,316 noncontrast chest CTs (2012–2017) from Duke University with 83 abnormality labels. We use its public subset (3,630 scans) as external validation, following CT-CLIP's framework \cite{hamamci2024ctrate} for consistency.

\noindent\textbf{Implementation details:}
\underline{Preprocessing}
All CT volumes are resampled to 1.5mm×1.5mm×3.0mm via trilinear interpolation. The intensity values are clipped to [-1000, 200] and then normalized to [-1,1]. Volumes are first padded/cropped to 240×240×120 and then randomly cropped to 192×192×96 during training, with 192×192×96 center crops for evaluation. 
\underline{Hyperparameters}
We use AdamW optimizer with a learning rate of 1e-4 for supervised pre-training and 1e-5 for CLIP training, with batch size set to 10 for both stages. For supervised pre-training we use equal weights for positive and negative samples. For CLIP training without L2 normalization, we multiply the loss by a coefficient of 0.1, as removing the L2 normalization increases the gradient scale.

\subsection{Experimental results}

\subsubsection{LLMs extract labels with a high precision}
Deepseek/Qwen annotated all 50,188 reports with the expected output format. Doubao exhibited two format errors, misclassifying the description "one millimetric nonspecific nodule in each lung" as (2) rather than the required binary value of presence (1) or absence (0). 

We evaluate the label quality by comparing the LLM-extracted labels to CT-RATE’s official BERT-extracted labels, which underwent quality checks with human annotations. The results are shown in Table \ref{tab:llm-metrics}. 
After analyzing the distribution of all abnormalities, we find "Medical material", "Atelectasis", and "Lung opacity" show high variance among labels extracted by different LLMs, we attribute this to definition ambiguity of these classes. For the remaining abnormality classes, all LLMs demonstrated robust performance in identifying condition presence and distinguishing between positive/negative mentions.

\begin{table}[t]
    \centering
    \caption{Quality comparison of labels extracted by different LLMs}
    \label{tab:llm-metrics}
    \begin{tabular}{lcccccc}
    \toprule
    Model & AUC & Accuracy & Precision & F1 & Sensitivity & Specificity \\
    \midrule
    Deepseek & 96.24 & 96.22 & 87.42 & 90.95 & 96.17 & 96.31 \\
    Qwen & 94.12 & 95.73 & 87.44 & 88.56 & 91.44 & 96.80 \\
    Doubao & 96.00 & 97.05 & 92.47 & 92.79 & 93.84 & 98.17 \\
    \bottomrule
    \end{tabular}
\end{table}

\noindent\textbf{LLM labeling enables large-scale supervised pre-training}
To assess the effectiveness of LLM-extracted labels, we train several 3D ResNet-18 models and compare their performance with models trained on CT-RATE's official labels. Remarkably, despite potential variations in label quality, models using LLM-extracted labels achieve performance comparable to those using the official labels. Furthermore, when merging labels from all three LLMs by averaging their predictions, the performance improves even further—surpassing the model trained on CT-RATE official labels.
This result is particularly significant considering the stark difference in annotation costs and effort.

\begin{table}[t]
    \centering
    \caption{Performance of vision encoder trained on different labels}
    \label{tab:performance-comparison-of-different-labels}
    \begin{tabular}{lcccccc}
    \toprule
    Model & AUC & Accuracy & Precision & F1 & Sensitivity & Specificity \\
    \midrule
    CT-RATE-official & 84.83 & 77.81 & 44.37 & 80.08 & 78.88 & 77.52 \\
    Deepseek         & 84.66 & 77.65 & 43.81 & 79.99 & 78.31 & 77.58 \\
    Qwen            & 84.18 & 77.84 & 44.14 & 80.01 & 77.13 & 78.19 \\
    Doubao          & 84.88 & 77.86 & 44.34 & 80.12 & 78.22 & 77.80 \\
    Merged          & 84.92 & 77.92 & 44.76 & 80.18 & 78.69 & 77.89 \\
    \bottomrule
    \end{tabular}
\end{table}

\begin{figure}[t]
    \centering
    \begin{subfigure}[b]{0.49\textwidth}
    \includegraphics[width=\textwidth]{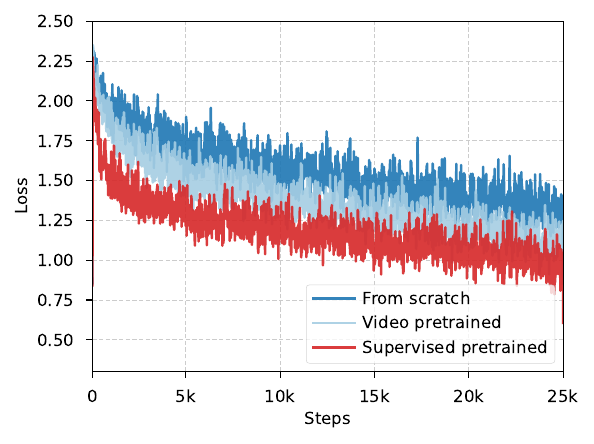}
    \caption{Loss curve with different vision encoder}
    \label{fig:clip-training-curve}
    \end{subfigure}
    \hfill
    \begin{subfigure}[b]{0.49\textwidth}
    \includegraphics[width=\textwidth]{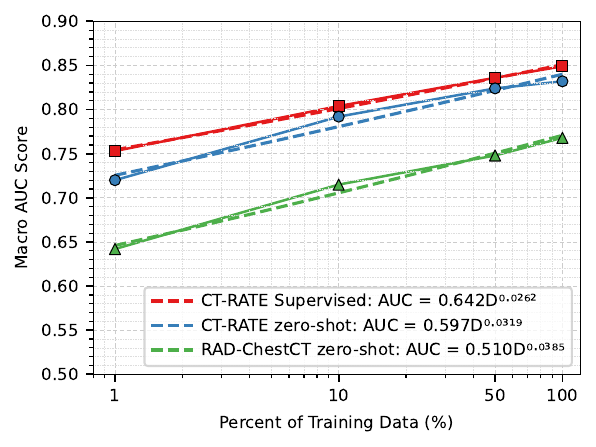}
    \caption{Data scaling law of our method}
    \label{fig:data-scaling-law}
    \end{subfigure}
    \caption{Supervised pretraining enhances vision-language alignment and scalability}
    \label{fig:combined}
\end{figure}

\begin{table*}[t]
    \centering
    \caption{Superivsed pre-training advance vision-language alignment}
    \label{tab:supervised-alignment}
    \begin{tabular}{lccccc}
    \toprule
    Model & \makecell{Sup.\\AUC} & \makecell{CT-RATE\\AUC} & \makecell{RAD-Chest\\AUC} & \makecell{Image-Image\\(MAP@5/10/50)} & \makecell{Report-Image\\(R@5/10/50/100)}   \\
    \midrule
    Base   & 84.9 & 79.0 & 69.1 & 70.2/59.8/51.8 & 5.5/9.4/29.2/41.7 \\
    Base+ASS            & 85.7 & 80.6 & 71.2 & 70.8/60.9/53.3 & 5.7/10.0/29.8/42.4 \\
    Base+ASS+LS            & 86.1 & 81.7 & 72.2 & 70.9/61.2/53.6 & 6.7/11.2/30.9/44.4 \\
    Base+ASS+LS$^\dagger$  & 86.1 & 83.8 & 77.3 & 71.3/61.5/53.7 & 10.0/16.3/39.0/52.2 \\
    \bottomrule
    \end{tabular}
    \begin{tablenotes}
        \small
        \item Sup.: Supervised; Base: 3D ResNet-18; ASS: auxiliary segmentation supervision; LS: label smoothing; $\dagger$: remove L2 normalization during CLIP training. For Base+ASS+LS$^\dagger$, the image-to-image retrieval is evaluated with L2 normalization.
    \end{tablenotes}
\end{table*}

\subsubsection{Supervised pre-training advances vision-language alignment}
Our experiments demonstrate a clear correlation between supervised pre-training performance and vision-language alignment capabilities. As shown in Table \ref{tab:supervised-alignment}, we observe that each enhancement to the supervised performance consistently leads to improvements across all vision-language evaluation metrics. In Figure \ref{fig:clip-training-curve}, we show that supervised pre-training considerably lowers the contrastive loss during CLIP training, indicating that the representation learned during supervised pre-training reduces the gap needing to be bridged in vision-language alignment. 

Notably, removing L2 normalization during CLIP training yields substantial gains, particularly in zero-shot diagnosis and report-image retrieval tasks. This improvement is unique to models initialized with supervised pre-trained weights, as similar modifications show minimal impact on randomly initialized or video pre-trained weights, demonstrating the importance of preserving the representation learned during supervised pre-training.

We analyze the data scaling law of our method, which is shown in Figure \ref{fig:data-scaling-law}. Our method exhabits a much larger coefficient compared to those reported in previous work \cite{shui2025largescale,blankemeier2024merlin}, indicating its exceptional data efficiency. 
While the exponential term is relatively smaller, we attribute this to the fact that our method surpasses their performance with only 10\% of the training data. And such a high performance slows down the rate of performance improvement.

\subsubsection{New SOTA performances with supervised pre-trained weights}

Building on the foundation of a strong supervised pre-trained model, we greatly advance the performance of vision-language alignment, establishing new state-of-the-art performance in zero-shot diagnosis, image-to-image retrieval and image-to-report cross-modal retrieval tasks.
As shown in Table \ref{tab:comparison}, our method achieves significant improvements over previous approaches, with an AUC of 83.8\% on CT-RATE (4.6\% absolute improvement over BrgSA) and 77.3\% on RAD-ChestCT (7.3\% improvement). This consistent performance gain across both internal and external validation sets demonstrates the robust generalization capability of our approach. Notably, the improvement margin is even larger on the external RAD-ChestCT dataset, suggesting that our supervised pre-training strategy helps learn more generalizable features.
The results in Table \ref{tab:merged_results} show that our method substantially improves both image-image and report-image retrieval tasks. 
It is worth noting that these improvements are achieved using a relatively lightweight 3D ResNet-18 architecture, while many compared methods use larger models or more complex architectures with ImageNet pre-trained weights or MAE pre-trained weights. This efficiency is particularly important for clinical applications where computational resources may be limited.

\begin{table*}[t]
    \centering
    \footnotesize
    \caption{Zero-shot abnormality diagnosis performance comparison across internal (CT-RATE) and external (RAD-ChestCT) validation benchmarks. The best performance is highlighted as bold.}
    \begin{tabular}{@{}l cccccccc@{}}
    \toprule
    Method & \multicolumn{4}{c}{CT-RATE} & \multicolumn{4}{c}{RAD-ChestCT} \\
    \cmidrule(lr){2-5} \cmidrule(lr){6-9}
          & AUC & ACC & F1 score & Precision & AUC & ACC & F1 score & Precision \\
    \midrule
    CT-CLIP & 73.1 & 66.8 & 70.7 & 32.3 & 62.9 & 59.5 & 64.2 & 33.6 \\
    BIUD \cite{cao2024bootstrapping} & 71.3 & 68.1 & 71.6 & 33.8 & 62.9 & 60.6 & 65.2 & 33.7 \\
    Merlin & 72.8 & 67.2 & 70.9 & 33.7 & 64.4 & 61.9 & 66.3 & 34.8 \\
    fLVM & 77.8 & 71.8 & 75.1 & 37.9 & 68.0 & 64.7 & 68.8 & 37.4 \\
    BrgSA & 79.2 & 73.3 & 76.2 & 38.5 & 70.0 & 65.5 & 69.3 & 39.1 \\
    Ours & \textbf{83.8} & \textbf{77.4} & \textbf{79.6} & \textbf{43.5} & \textbf{77.3} & \textbf{71.9} & \textbf{74.8} & \textbf{45.0} \\
    \bottomrule
    \end{tabular}
    \label{tab:comparison}
\end{table*}

\begin{table}[t]
    \centering
    \setlength{\tabcolsep}{12pt} %
    \footnotesize
    \caption{Performance comparison for image-image and image-report retrieval tasks on CT-RATE dataset. The best performance is highlighted as bold.}
    \begin{tabular}{@{}l >{\centering\arraybackslash}p{3.8cm} >{\centering\arraybackslash}p{4cm}@{}}
    \toprule
    Method & 
    \makecell{Image-Image Retrieval \\ (MAP@5/10/50)} & 
    \makecell{Report-Image Retrieval \\ (Recall@5/10/50/100)} \\
    \midrule
    VocabFine  & 68.3/57.2/48.8        & 0.1/0.6/2.3/2.0      \\
    ClassFine  & 67.9/56.8/48.5        & --/--/--/--          \\
    CT-CLIP    & 68.3/57.2/48.9        & 2.9/5.0/18.0/28.7    \\
    Merlin     & 62.6/51.3/43.9        & 1.5/2.7/7.7/12.7     \\
    BrgSA      & 69.2/58.5/50.5        & 5.8/10.1/28.6/42.0  \\
    Ours       & \textbf{71.3/61.5/53.7}        & \textbf{10.0/16.3/39.0/52.2}  \\
    \bottomrule
    \end{tabular}
    \label{tab:merged_results}
\end{table}

\section{Conclusion}
\label{sec:conclusion}
We introduce an efficient framework that demonstrates the potential of constructing large-scale annotated datasets with LLMs and using them to train a performant supervised learning models at a very low cost, 
which further advances vision-language alignment, leading to new state-of-the-art performance in zero-shot diagnosis, image-to-image, and image-to-report cross-modal retrieval with a 3D ResNet-18 through a vanilla CLIP training pipeline. 
The success of our method suggests the potential of leveraging LLMs to faciliate more performant and scalable medical AI systems.



\begin{credits}
\subsubsection{\ackname} Supported by Natural Science Foundation of China under Grant 62271465, National Key R\&D Program of China under Grant 2025YFC3408300, and Suzhou Basic Research Program under Grant SYG202338.

\subsubsection{\discintname}
The authors have no competing interests to declare that are relevant to the content of this article.
\end{credits}

\bibliographystyle{splncs04}
\bibliography{reference}

\begin{thebibliography}{10}
\providecommand{\url}[1]{\texttt{#1}}
\providecommand{\urlprefix}{URL }
\providecommand{\doi}[1]{https://doi.org/#1}

\bibitem{alayrac2022flamingo}
Alayrac, J.B., Donahue, J., Luc, P., Miech, A., Barr, I., Hasson, Y., Lenc, K., Mensch, A., Millican, K., Reynolds, M., et~al.: Flamingo: a visual language model for few-shot learning. Advances in neural information processing systems  \textbf{35},  23716--23736 (2022)

\bibitem{ardila2019end}
Ardila, D., Kiraly, A.P., Bharadwaj, S., Choi, B., Reicher, J.J., Peng, L., Tse, D., Etemadi, M., Ye, W., Corrado, G., et~al.: End-to-end lung cancer screening with three-dimensional deep learning on low-dose chest computed tomography. Nature medicine  \textbf{25}(6),  954--961 (2019)

\bibitem{bai2023qwen}
Bai, J., Bai, S., Chu, Y., Cui, Z., Dang, K., Deng, X., Fan, Y., Ge, W., Han, Y., Huang, F., et~al.: Qwen technical report. arXiv preprint arXiv:2309.16609  (2023)

\bibitem{blankemeier2024merlin}
Blankemeier, L., Cohen, J.P., Kumar, A., Van~Veen, D., Gardezi, S.J.S., Paschali, M., Chen, Z., Delbrouck, J.B., Reis, E., Truyts, C., et~al.: Merlin: A vision language foundation model for 3d computed tomography. Research Square pp. rs--3 (2024)

\bibitem{boecking2022cxrbert}
Boecking, B., Usuyama, N., Bannur, S., Castro, D.C., Schwaighofer, A., Hyland, S., Wetscherek, M., Naumann, T., Nori, A., Alvarez-Valle, J., et~al.: Making the most of text semantics to improve biomedical vision--language processing. In: European conference on computer vision. pp. 1--21. Springer (2022)

\bibitem{doubao2023}
ByteDance: Doubao: A large language model by bytedance. \url{https://www.doubao.com} (2023), accessed: June 2024

\bibitem{cao2023large}
Cao, K., Xia, Y., Yao, J., Han, X., Lambert, L., Zhang, T., Tang, W., Jin, G., Jiang, H., Fang, X., et~al.: Large-scale pancreatic cancer detection via non-contrast ct and deep learning. Nature medicine  \textbf{29}(12),  3033--3043 (2023)

\bibitem{cao2024bootstrapping}
Cao, W., Zhang, J., Xia, Y., Mok, T.C., Li, Z., Ye, X., Lu, L., Zheng, J., Tang, Y., Zhang, L.: Bootstrapping chest ct image understanding by distilling knowledge from x-ray expert models. In: Proceedings of the IEEE/CVF Conference on Computer Vision and Pattern Recognition. pp. 11238--11247 (2024)

\bibitem{cid2024development}
Cid, Y.D., Macpherson, M., Gervais-Andre, L., Zhu, Y., Franco, G., Santeramo, R., Lim, C., Selby, I., Muthuswamy, K., Amlani, A., et~al.: Development and validation of open-source deep neural networks for comprehensive chest x-ray reading: a retrospective, multicentre study. The Lancet Digital Health  \textbf{6}(1),  e44--e57 (2024)

\bibitem{dai2024deepseekmoe}
Dai, D., Deng, C., Zhao, C., Xu, R., Gao, H., Chen, D., Li, J., Zeng, W., Yu, X., Wu, Y., et~al.: Deepseekmoe: Towards ultimate expert specialization in mixture-of-experts language models. arXiv preprint arXiv:2401.06066  (2024)

\bibitem{draelos2021radchestct}
Draelos, R.L., Dov, D., Mazurowski, M.A., Lo, J.Y., Henao, R., Rubin, G.D., Carin, L.: Machine-learning-based multiple abnormality prediction with large-scale chest computed tomography volumes. Medical image analysis  \textbf{67},  101857 (2021)

\bibitem{hamamci2024ctrate}
Hamamci, I.E., Er, S., Almas, F., Simsek, A.G., Esirgun, S.N., Dogan, I., Dasdelen, M.F., Wittmann, B., Simsar, E., Simsar, M., et~al.: A foundation model utilizing chest ct volumes and radiology reports for supervised-level zero-shot detection of abnormalities. CoRR  (2024)

\bibitem{kay2017kinetics}
Kay, W., Carreira, J., Simonyan, K., Zhang, B., Hillier, C., Vijayanarasimhan, S., Viola, F., Green, T., Back, T., Natsev, P., et~al.: The kinetics human action video dataset. arXiv preprint arXiv:1705.06950  (2017)

\bibitem{lai2025bridged}
Lai, H., Jiang, Z., Yao, Q., Wang, R., He, Z., Tao, X., Wei, W., Lv, W., Zhou, S.K.: Bridged semantic alignment for zero-shot 3d medical image diagnosis. arXiv preprint arXiv:2501.03565  (2025)

\bibitem{li2023blip2}
Li, J., Li, D., Savarese, S., Hoi, S.: Blip-2: Bootstrapping language-image pre-training with frozen image encoders and large language models. In: International conference on machine learning. pp. 19730--19742. PMLR (2023)

\bibitem{li2025supervisionscaling}
Li, Y., Ming, S., Lai, H., Tang, F., Wei, W., Zhou, S.K.: Scaling supervision for free: Leveraging universal segmentation models for enhanced medical image diagnosis. In: Submitted to Medical Imaging with Deep Learning (2025), \url{https://openreview.net/forum?id=SpHsR20XjU}, under review

\bibitem{liu2024visualinstructtuning}
Liu, H., Li, C., Wu, Q., Lee, Y.J.: Visual instruction tuning. Advances in neural information processing systems  \textbf{36} (2024)

\bibitem{park2024automated}
Park, R.Y., Windsor, R., Jamaludin, A., Zisserman, A.: Automated spinal mri labelling from reports using a large language model. In: International Conference on Medical Image Computing and Computer-Assisted Intervention. pp. 101--111. Springer (2024)

\bibitem{radford2021clip}
Radford, A., Kim, J.W., Hallacy, C., Ramesh, A., Goh, G., Agarwal, S., Sastry, G., Askell, A., Mishkin, P., Clark, J., et~al.: Learning transferable visual models from natural language supervision. In: International conference on machine learning. pp. 8748--8763. PmLR (2021)

\bibitem{shui2025largescale}
Shui, Z., Zhang, J., Cao, W., Wang, S., Guo, R., Lu, L., Zhang, L., Liang, T., Yang, L., Ye, X., Zhang, Q.: Large-scale and fine-grained vision-language pre-training for enhanced ct image understanding. In: The Thirteenth International Conference on Learning Representations (2025), \url{https://openreview.net/forum?id=nYpPAT4L3D}

\bibitem{smith2012use}
Smith-Bindman, R., Miglioretti, D.L., Johnson, E., Lee, C., Feigelson, H.S., Flynn, M., Greenlee, R.T., Kruger, R.L., Hornbrook, M.C., Roblin, D., et~al.: Use of diagnostic imaging studies and associated radiation exposure for patients enrolled in large integrated health care systems, 1996-2010. Jama  \textbf{307}(22),  2400--2409 (2012)

\bibitem{szegedy2016labelsmoothing}
Szegedy, C., Vanhoucke, V., Ioffe, S., Shlens, J., Wojna, Z.: Rethinking the inception architecture for computer vision. In: Proceedings of the IEEE conference on computer vision and pattern recognition. pp. 2818--2826 (2016)

\bibitem{tiu2022chexzero}
Tiu, E., Talius, E., Patel, P., Langlotz, C.P., Ng, A.Y., Rajpurkar, P.: Expert-level detection of pathologies from unannotated chest x-ray images via self-supervised learning. Nature Biomedical Engineering  \textbf{6}(12),  1399--1406 (2022)

\bibitem{wang2022development}
Wang, C., Ma, J., Zhang, S., Shao, J., Wang, Y., Zhou, H.Y., Song, L., Zheng, J., Yu, Y., Li, W.: Development and validation of an abnormality-derived deep-learning diagnostic system for major respiratory diseases. NPJ Digital Medicine  \textbf{5}(1), ~124 (2022)

\bibitem{wang2024screening}
Wang, Y.R., Yang, K., Wen, Y., Wang, P., Hu, Y., Lai, Y., Wang, Y., Zhao, K., Tang, S., Zhang, A., et~al.: Screening and diagnosis of cardiovascular disease using artificial intelligence-enabled cardiac magnetic resonance imaging. Nature Medicine pp. 1--10 (2024)

\bibitem{zhai2022lit}
Zhai, X., Wang, X., Mustafa, B., Steiner, A., Keysers, D., Kolesnikov, A., Beyer, L.: Lit: Zero-shot transfer with locked-image text tuning. In: Proceedings of the IEEE/CVF conference on computer vision and pattern recognition. pp. 18123--18133 (2022)

\bibitem{zhang2023KAD}
Zhang, X., Wu, C., Zhang, Y., Xie, W., Wang, Y.: Knowledge-enhanced visual-language pre-training on chest radiology images. Nature Communications  \textbf{14}(1), ~4542 (2023)

\end{thebibliography}

\end{document}